\documentclass[10pt]{article} 

\usepackage[preprint]{rlj} 

\usepackage{amssymb}            
\usepackage{mathtools}          
\usepackage{mathrsfs}           
\usepackage{graphicx}           
\usepackage{subcaption}         
\usepackage[space]{grffile}     
\usepackage{url}                
\usepackage{lipsum}             
\usepackage{xurl}
\usepackage{placeins}
\usepackage{algorithm}
\usepackage{algpseudocode}
\usepackage{afterpage}
\usepackage{dsfont}
\usepackage{pifont}
\usepackage{amsthm}
\usepackage{wrapfig}

\newcommand{\mT}{\mathcal{T}}
\newcommand{\mM}{\mathcal{M}}

\newcommand{\mX}{\mathcal{X}}
\newcommand{\mU}{\mathcal{U}}

\newcommand{\mH}{\mathcal{H}}

\newcommand{\cmark}{\textcolor{green}{\ding{51}}} 
\newcommand{\xmark}{\textcolor{red}{\ding{55}}}            

\DeclareMathOperator*{\argmax}{arg\,max}

\newcommand{\phigrid}{\mH_\mathrm{grid}}
\newcommand{\phiroom}{\mH_\mathrm{room}}

\newcommand{\Vstar}{V_{\mH^\tau}^*}
\newcommand{\pistar}{\pi_{\mH^\tau}^*}

\newtheorem{proposition}{Proposition}

\algrenewcommand{\Return}{\State\algorithmicreturn~}
\title{Learning with Expert Abstractions for Efficient Multi-Task Continuous Control}
\setrunningtitle{Learning with Expert Abstractions}
\author{Jeff Jewett\textsuperscript{1}, Sandhya Saisubramanian\textsuperscript{1}}

\emails{\{jewettje, sandhya.sai\}@oregonstate.edu}

\affiliations{
$^{1}$\textbf{School of EECS, Oregon State University, Corvallis, OR, United States}
}

\keywords{Hierarchical RL, Abstractions, Multi-task, Zero-shot, Continuous Control, Planning }

\begin{document}

\maketitle  

\begin{abstract}
Decision-making in complex, continuous multi-task environments is often hindered by the difficulty of obtaining accurate models for planning and the inefficiency of learning purely from trial and error. While precise environment dynamics may be hard to specify, human experts can often provide high-fidelity abstractions that capture the essential high-level structure of a task and user preferences in the target environment. Existing hierarchical approaches often target discrete settings and do not generalize across tasks. We propose a hierarchical reinforcement learning approach that addresses these limitations by dynamically planning over the expert-specified abstraction to generate subgoals to learn a goal-conditioned policy. To overcome the challenges of learning under sparse rewards, we shape the reward based on the optimal state value in the abstract model. This structured decision-making process enhances sample efficiency and facilitates zero-shot generalization. Our empirical evaluation on a suite of procedurally generated continuous control environments demonstrates that our approach outperforms existing hierarchical reinforcement learning methods in terms of sample efficiency, task completion rate, scalability to complex tasks, and generalization to novel scenarios.\footnote{Code for our method and experiments is at \url{https://github.com/Intelligent-Reliable-Autonomous-Systems/gcrs-expert-abstractions}}

\end{abstract}

\section{Introduction}
\label{sec:introduction}
Reinforcement learning (RL) has demonstrated promising results in many domains such as robotics and game playing~\citep{arulkumaran_deep_2017}. Despite the significant advances, scaling RL to real-world continuous control environments remains a significant challenge due to high sample complexity, long horizons, sparse rewards, and the need to generalize across tasks~\citep{cobbe_quantifying_2019}.
A promising alternative to direct policy learning is model-based RL, where an agent leverages an explicit model of the environment to plan efficiently~\citep{luo_survey_2024}. However, obtaining an accurate dynamics model for real-world tasks is often infeasible. Even when models can be learned, they tend to be inaccurate in high-dimensional settings due to compounding error~\citep{lambert_investigating_2022}. In the absence of a perfect model for planning, structured priors such as abstractions encode high-level user preferences and help improve learning efficiency while maintaining adaptability. 

\begin{figure}[t]
    \includegraphics[width=0.95\textwidth]{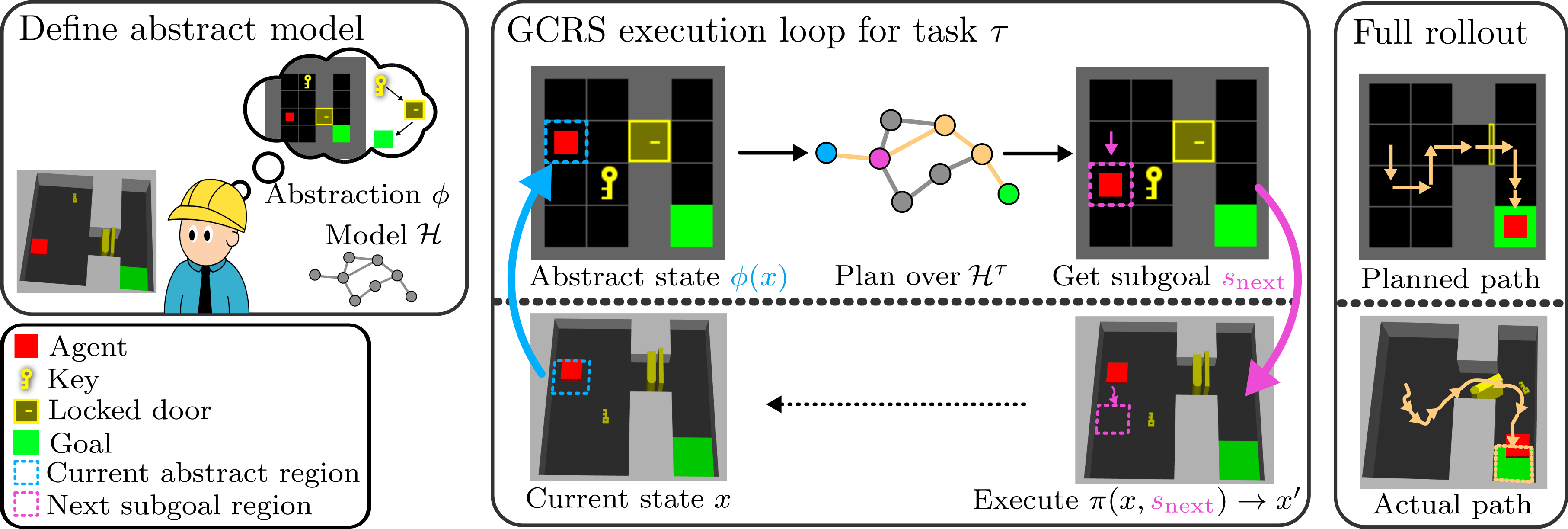}
        \caption{Overview of our proposed solution approach. An expert encodes their domain knowledge as an abstract model $\mH$ with a mapping $\phi$ from continuous to abstract states. At each step for a given task $\tau$, map the current state $x$ to abstract state $\phi(x)$. Plan the best path with $\mH^\tau$ and feed $s_\mathrm{next}$ into the policy to get next state $x'$. The agent is rewarded for reaching higher value abstract states.}
        \label{fig:overview}
\end{figure}

While precise low-level models are difficult to obtain and specify, human experts can often provide high-fidelity, high-level abstractions that capture the essential structure of a task in the target environment~\citep{yu_reinforcement_2023}. These abstractions are also a tool for the user to encode their preferences that are not captured by the sparse reward function, thereby effectively guiding the agent in exploring the environment and learning a low-level control policy. For example in a continuous version of the DoorKey domain~\citep{chevalier-boisvert_minigrid_2023}, the agent must pick up a key in one room to unlock a door leading to a goal location. This is a sparse reward setting where the agent must navigate hundreds of steps to reach the goal with no intermediate reward. In addition, the key and door locations are randomly generated, making it a challenging learning problem.
However, it is relatively straightforward for a human to design abstractions for this domain, such as a grid discretization shown in Figure \ref{fig:overview}, which can be used to generate meaningful subgoals (e.g., ``move to'', ``pick up key'', ``open door'') for learning in the continuous setting.

There is a rich literature on using abstractions for learning~\citep{hutsebaut-buysse_hierarchical_2022}. The abstractions are used to generate high-level policies in hierarchical RL (HRL), such as options or subgoals in goal-conditioned RL. Options are macro-actions that can be reused across tasks, but they require learning an individual skill policy for each macro-action.
Deriving a single policy by combining different options is non-trivial due to the differences in their termination and initiation distributions~\citep{jothimurugan_abstract_2021}. Goal-conditioned RL (GCRL) enables agents to learn generalizable skills as a single policy~\citep{schaul_universal_2015}. However, the learned goal representations struggle to generalize to new environments~\citep{hutsebaut-buysse_hierarchical_2022}.
Some hierarchical methods have incorporated structured expert knowledge to guide learning, but they largely target discrete gridworld settings~\citep{sun_program_2019, zhao_consciousness-inspired_2023} and require additional training to adapt to new tasks~\citep{illanes_symbolic_2020,kim_landmark-guided_2021}.

We present a hierarchical RL framework, Goal-Conditioned Reward Shaping (GCRS), that addresses the limitations of existing HRL methods by leveraging expert-defined abstractions as high-level models for 
reward shaping and subgoal selection in continuous control tasks. Our approach dynamically plans over an expert-defined abstraction to generate adaptive subgoals for a goal-conditioned policy (outlined in Figure~\ref{fig:overview}). This enables sample efficient learning by focusing on task-relevant exploration, determined by the abstraction, and zero-shot generalization~\citep{kirk_survey_2023} to previously unseen scenarios. Our extensive empirical evaluation on a suite of procedurally generated continuous control environments demonstrates that our approach is sample efficient, scalable, and generalizes to novel scenarios, compared to the existing HRL and GCRL methods.

\section{Related Works}
\label{sec:related_works}

\noindent \textbf{Hierarchical RL} Hierarchical RL (HRL) divides long, complex problems into manageable sub-problems. HRL is built on temporal abstractions, treating multi-step behaviors as abstracted actions, enabling composition of larger building blocks~\citep{hutsebaut-buysse_hierarchical_2022}. Temporal abstractions are constructed through two main approaches: the Options framework~\citep{sutton_between_1999} and Goal-Conditioned RL (GCRL)~\citep{schaul_universal_2015}. Options are individual skill policies that can be initiated and executed until a termination state. 
However, each skill is learned independently as a separate policy, which ignores potential overlap from shared dynamics and increases the training time with the number of skills~\citep{hutsebaut-buysse_hierarchical_2022}. 
GCRL instead focuses on learning a single controller policy, where experiences for different subgoals contribute to a shared representation, allowing the policy to scale to more subgoals efficiently.
A hierarchical structure can be imposed with a manager policy that selects subgoals to guide the agent to the final goal.

\noindent \textbf{Learning with abstractions} 
High-level policies are significantly more efficient with abstract representations that shrink the problem down to salient features. While these abstractions can be learned autonomously through observations~\citep{bacon_option-critic_2017}, domain experts often define structured abstractions that guide efficient learning and generalization to new environments~\citep{yu_reinforcement_2023}.

Several approaches leverage expert abstractions, often in the form of feature selection of agent position or symbolic representations, to guide hierarchical decision-making. \citet{kim_landmark-guided_2021} learn a graph of reachable agent positions and a goal-conditioned policy to navigate through these waypoints.
\citet{jothimurugan_abstract_2021} use expert-defined subgoal regions which are treated as nodes in an abstract MDP to learn a skill policy for each abstract transition.
Symbolic abstractions are also commonly used to learn a skill policy for each symbolic action in the plan~\citep{lyu_sdrl_2019,yu_reinforcement_2023}. 
Plan-based reward shaping leverages symbolic planning to automatically construct a potential-based reward function~\citep{grzes_plan-based_2008, canonaco_sample_2024}.
Expert abstractions have also been used to enable zero-shot generalization, adapting to new tasks without additional training. \citet{zhao_consciousness-inspired_2023} dynamically construct and plan over a graph of abstract positions, but are restricted to discrete settings. Attribute Planner uses an exploration policy to learn relationships between abstract attributes, then dynamically plans a path based on empirical success~\citep{zhang_composable_2018}. However, planning can be performed only with abstract states seen in training. Additionally, these methods \emph{do not} consider long-horizon, continuous-action tasks. \citet{vaezipoor_ltl2action_2021} demonstrate a goal-conditioned policy with continuous actions that can generalize to new temporal logic specifications, but assume the subgoals are given and do not address reward sparsity.

Our approach builds on this foundation by planning over an expert-defined abstraction to decompose each task into a sequence of subgoals. Unlike prior works that limit themselves to discrete settings with short horizons, our approach successfully completes  continuous-action tasks requiring hundreds of steps. A key limitation of Options framework is that it requires learning a policy for each skill, scaling poorly in larger abstractions. We overcome this drawback by learning a single goal-conditioned control policy. The learned controller can follow the guidance of a high-level planner to adapt to new environments zero-shot. Table \ref{table:comparison} summarizes the key differences between our approach and prior HRL methods that leverage expert abstractions.

\begin{table}[t]
\centering
\small 
\begin{tabular}{l c c c c}
\textbf{Approach}
             & \shortstack{\textbf{Generate} \\ \textbf{Subtasks}} 
             & \shortstack{\textbf{Continuous} \\ \textbf{Actions}} 
             & \shortstack{\textbf{Single Learned} \\ \textbf{Controller}} 
             & \shortstack{\textbf{Zero-Shot} \\ \textbf{Generalization}} \\
\hline \\
\citet{grzes_plan-based_2008} & \cmark & \xmark & \cmark & \xmark \\
\citet{zhang_composable_2018}        & \cmark & \xmark & \xmark & \cmark \\
\citet{sun_program_2019}          & \xmark & \xmark & \cmark & \cmark \\
\citet{lyu_sdrl_2019}           & \cmark & \xmark & \xmark & \xmark \\
\citet{illanes_symbolic_2020}     & \cmark & \xmark & \xmark & \xmark \\
\citet{kim_landmark-guided_2021}          & \cmark & \cmark & \cmark & \xmark \\
\citet{jothimurugan_abstract_2021} & \cmark & \cmark & \xmark & \xmark \\
\citet{vaezipoor_ltl2action_2021}    & \xmark & \cmark & \cmark & \cmark \\
\citet{zhao_consciousness-inspired_2023}  & \cmark & \xmark & \cmark & \cmark \\
\citet{canonaco_sample_2024} & \cmark & \xmark & \cmark & \xmark \\
Ours                   & \cmark & \cmark & \cmark & \cmark
\end{tabular}
\caption{A comparison of hierarchical RL methods that use some degree of expert-informed abstraction. Generate Subtasks indicates that a method can decompose tasks into subtasks. A Single Learned Controller means only one low-level policy is learned. Zero-Shot Generalization is whether the learned policy can adapt to changes in the environment without additional training.}
\label{table:comparison}
\end{table}

\section{Problem Setting}
\label{sec:problem_setting}

Many real-world applications require an agent to complete a set of tasks $\mT$, each defined by reaching a goal from a given initial state. Consider a task-family $\mM \doteq \{ \mM^\tau \mid \tau \in \mT \}$ of Markov Decision Processes (MDPs), with $\mM^\tau \doteq \langle \mX, \mU, p, G^\tau, F^\tau, R^\tau, \mu_0^\tau, \gamma \rangle $ where $\mX$ and $\mU$ are continuous state and action spaces, shared across all tasks. $p$ is the dynamics function $p(x, u, x') \doteq \Pr(x_{t+1}=x' \mid x_t = x, u_t = u)$. $G^\tau \subset \mX$ and $F^\tau \subset \mX$ are the set of terminating goal and failure states respectively. $R^\tau$ is the sparse task reward function $R^\tau(s,a,s') \doteq \mathds{1}[s' \in G^\tau]$ with a discount factor $\gamma$. The task initial state distribution is denoted by $\mu_0^\tau$. The objective is to find a policy $\pi : \mX \times \mT \to \Delta(\mU)$ that can select actions given a task context to maximize expected return over a task distribution $\mu_\mT$: $J(\pi) \doteq \mathbb{E}_{\tau \sim \mu_\mT} \mathbb{E}_{x_0 \sim \mu_0^\tau} \mathbb{E} \left[\sum_t \gamma^t R^\tau(x, u, x') | x_0, \pi \right]$. 

Similar to \citet{illanes_symbolic_2020, vaezipoor_ltl2action_2021, jothimurugan_abstract_2021}, we assume a domain expert can provide an \emph{abstract model} $\mH^\tau\!=\!\langle S, A, T, R_{\mH}^\tau\rangle$ for each task $\tau$ that captures some high-level features of the ground environment $\mM^\tau$. The states $S$ in the abstract MDP $\mH^\tau$ are related to the continuous states in $\mM^\tau$ by a \emph{state abstraction} $\phi : \mX \to S$ that is assumed to be available. 
$A$ denotes the set of macro-actions. 
We model the action outcomes in the abstract model as a deterministic function $T\!:\!S \times A \to S$ but it is straightforward to extend our approach to support stochastic transitions in $\mH^\tau$. 
The task-specific reward function $R_\mH^\tau$ incentivizes the expert's desired behavior.
For each $\mH^\tau$, the optimal state values  $V_{H^\tau}^{*}$ and optimal policy $\pi_{H^\tau}^{*}$ are calculated as, 
\begin{align*}
    V_{H^\tau}^{*}(s) \doteq& \max_{a \in A(s)} [R_\mH^\tau(s,a) + V_{H^\tau}^{*}(T(s,a))], \quad \forall s \in S \\
    \pi_{H^\tau}^{*}(s) \doteq& \argmax_{a \in A(s)} [R_\mH^\tau(s,a) + V_{H^\tau}^{*}(T(s,a))],  \quad \forall s \in S.
\end{align*}
An abstract model provided by a domain expert is typically significantly easier to solve than the ground MDP. The state and action spaces have reduced dimensionality and are often discrete. $V_{H^\tau}^{*}$ and $\pi_{H^\tau}^{*}$ can be solved with any standard method, such as a search algorithm or value iteration. Even if $V_{H^\tau}^{*}$ and $\pi_{H^\tau}^{*}$ can only be approximated, the approximations can be regarded as optimal solutions to a different abstract model. The main challenge is how to effectively use $V_{H^\tau}^{*}$ and $\pi_{H^\tau}^{*}$ to reduce the difficulty of solving $\mM^\tau$ and even solve new tasks zero-shot.

\section{Learning Goal-Conditioned Controller with Reward Shaping}
\label{sec:method}
We present Goal-Conditioned Reward Shaping (GCRS) to learn a continuous-control policy that uses the guidance of a high-level model to adapt to new tasks. The learned control policy is conditioned on subgoals generated by the high-level planner. To effectively learn under sparse rewards, we use a potential-based reward function, based on the plan computed using the high-level model. 

\paragraph{Plan-based reward shaping}
Learning to successfully complete
sparse-reward, long-horizon tasks is difficult since the agent does not receive immediate feedback on its actions. Reward shaping is the use of auxiliary rewards as a heuristic to guide the learning process towards optimal behavior. A common form of reward shaping is the potential-based reward shaping~\citep{ng_policy_1999} that uses a state potential function $\phi$ to 
produce a shaped reward of the form $R_\mathrm{shaped}(x, u, x')=R_{original}(x,u, x') + \gamma \Phi(x') - \Phi(x)$. 
Similar to \citet{canonaco_sample_2024} who utilize an abstraction for reward shaping, we design our potential function for a given task $\tau$ based on the optimal state value in the abstract model, $\Phi^\tau(x) = V_{H^\tau}^{*}(\phi(x))$.

\paragraph{Goal-conditioned controller}
A successful policy must respond to structural variations in the distribution of environments. However, it is challenging to transfer learned representations to new environments \citep{cobbe_quantifying_2019}. One promising approach to enable zero-shot generalization is to decompose tasks into subgoals~\citep{zhang_composable_2018}. In our case, a task $\tau$ can be decomposed into subgoals based on $\pistar$. We leverage this task decomposition and use a plan-based reward shaping to learn a goal-conditioned policy $\pi\!:\!\mX\!\times S \times \mT\!\to\!\Delta(U)$. The policy takes as input the current continuous state $x \in \mX$, a subgoal $s_\mathrm{next}\in S$, and the task context $\tau \in \mT$ and outputs a distribution of actions. The subgoal for $\tau$ is generated by abstracting $x$ to $\phi(x)$ and calculating the next abstract state based on the macro-action $\pi_{\mH^\tau}^*(\phi(x))$. If $\phi(x)$ is already a terminal abstract goal or failure state, then it is used as the subgoal:
\begin{align}
    \mathop{subgoal}(x, \tau) =& 
        \begin{cases}
        T(\phi(x), \pi_{\mH^\tau}^*(\phi(x))) & \text{if } \phi(x) \text{ is nonterminal}\\
        \phi(x) & \text{otherwise.}
        \end{cases} \label{eq:next_abstract}
\end{align}

Algorithm \ref{alg:rl_amdp} describes the full process to learn $\pi$ online. A task is sampled in each episode. The corresponding abstract task $\mH^\tau$ is solved optimally using any method, such as a search algorithm or value iteration (Line \ref{line:solve_h_tau}). In each step of the episode, a subgoal is queried according to Equation \ref{eq:next_abstract} (Line \ref{line:get_subgoal}). An action from the policy conditioned on that subgoal is executed following $\pi$ (Lines \ref{line:get_action}-\ref{line:execute}). The value of the resulting abstract state forms a difference in potential, providing a shaped reward (Lines \ref{line:phi_prev}-\ref{line:rew_shaped}). Finally, the agent updates its policy $\pi$ based on the experience (Line~\ref{line:update_policy}). In our experiments we use Robust Policy Optimization~\citep{rahman_robust_2022}, but any learning algorithm can be used to update $\pi$. When a task is completed, failed, or times out, the current episode terminates and a new task is sampled. We boost efficiency by batching and performing lazy planning over $\mH^\tau$, triggering replanning when new abstract states are encountered.

\begin{algorithm}[t]
\caption{Goal-Conditioned Reward Shaping}
\label{alg:rl_amdp}
\begin{algorithmic}[1]
\Require MDP family \( \mathcal{M} \), Abstract MDP family $\mH$, task distribution $ \mu_{\mathcal{T}}$, \# training steps $N_\mathrm{steps}$
\State Initialize policy \( \pi \)
\State \( \text{terminate} \gets \mathop{true} \)
\For{\( i = 0 \) to \( N_\mathrm{steps} \)}
    \If{\text{terminate}}
        \State Sample task \( \tau \sim \mu_{\mathcal{T}} \)
        \State \( x \gets \mathop{reset}(\mathcal{M}^\tau) \)
        \State \label{line:solve_h_tau} \( \Vstar,\, \pistar \gets \text{solve } \mH^\tau \)
    \EndIf
    
    \State \label{line:get_subgoal} \( s_\mathrm{next} \gets \mathop{subgoal}(x, \tau) \)
    \State \label{line:get_action} \( u \gets \pi(x, s_\mathrm{next}, \tau) \)
    \State \label{line:execute} \( x', r, \text{terminate} \gets \text{execute } u \text{ in } \mathcal{M}^\tau \)
    \State \label{line:phi_prev} \( \Phi_\text{prev} \gets V_{\mathcal{H}^\tau}^*(\phi(x)) \)
    \State \label{line:phi_next} \( \Phi_\text{next} \gets V_{\mathcal{H}^\tau}^*(\phi(x')) \) 
    \State \label{line:rew_shaped} \( r_\text{shaped} \gets r + \gamma \Phi_\text{next} - \Phi_\text{prev} \)
    \State \label{line:update_policy} Update policy \( \pi \) based on experience \( (x, s_\mathrm{current}, \tau, u, r_\text{shaped}) \) using any learning algorithm
    \State \( x \gets x'\)
\EndFor
\Return $\pi$
\end{algorithmic}
\end{algorithm}

GCRS learns a policy that is conditioned on subgoals generated by $\mH^\tau$ for any task $\tau$. Imposing hierarchical constraints can induce sub-optimality~\citep{dietterich_hierarchical_2000}. However, we show that this conditioning does not affect the optimal policy for $\mM$.

\begin{proposition}[Optimality] Let $\pi' : \mX \times S \times \mT \to \Delta(U)$ be the policy learned by Algorithm \ref{alg:rl_amdp} as $N_\mathrm{steps} \to \infty$. Suppose the policy update on Line~\ref{line:update_policy} is an RL procedure that converges to an optimal policy under its usual assumptions. Then $\pi(x,\tau) \doteq \pi'(x, \mathop{subgoal}(x,\tau), \tau)$ is optimal for the objective $\max_\pi J(\pi)$.
\end{proposition}

Proof is presented in Appendix \ref{sec:appendix1}. Intuitively, the abstract subgoal is determined by the continuous state and the task, and the dynamics are unaffected. Additionally, the use of PBRS does not change the optimal policy \citep{canonaco_sample_2024}. Therefore in the limit, the bias of even very inaccurate abstract models would be overcome. In the short term, however, we expect GCRS' use of expert knowledge can be beneficial, which we show experimentally in the next section.

\section{Experiments}
\label{sec:experiments}

We evaluate GCRS on a collection of continuous navigation and object manipulation tasks. The results are compared with two abstract models that capture high-level environment features, with different levels of detail. We test how well GCRS can utilize each abstraction to (1) learn sample efficiently, (2) complete tasks as difficulty and reward-sparsity of an environment increases, and (3) adapt zero-shot to new tasks and environment configurations.

\paragraph{Implementation and Baselines}

We train GCRS online with Robust Policy Optimization~\citep{rahman_robust_2022}. We solve the abstractions as shortest path planning with Djikstra's algorithm, using the negative of the path cost as the potential function and replanning when the agent deviates.

We compare GCRS against four HRL methods that utilize some degree of expert-provided abstraction and support continuous environments: (1) Plan-based reward shaping (Plan-RS) uses the reward shaping on a flat policy without subgoals; (2) LTL2Action~\citep{vaezipoor_ltl2action_2021} learns to satisfy abstract LTL specifications with a goal-conditioned policy, with subgoals generated by the same Djikstra planner as GCRS for a fair comparison; (3) AAVI~\citep{jothimurugan_abstract_2021} learns a separate policy for each macro-action; and (4) HIGL~\citep{kim_landmark-guided_2021} is a GCRL method that creates a graph of continuous position landmarks as the agent explores. The landmark graph is static at test time and does not account for the environment's changing structure. 

\paragraph{Environments}
Our evaluations use Minigrid environments~\citep{chevalier-boisvert_minigrid_2023} as the abstract models that guide learning in CocoGrid (Continuous Control Minigrid), our continuous extension of Minigrid. Minigrid is a discrete gridworld environment designed for goal-oriented 
tasks. It is straightforward to scale the environment complexity in size, number of objects, location and number of walls, and tasks. CocoGrid extends Minigrid to support physical scenes, benefiting from the customizability of Minigrid. The agent is a 2-DoF rectangular point mass equipped with a magnet action to drag a nearby object. 

We evaluate the performance of different HRL approaches on multiple task configurations that involve hundreds of steps with a sparse binary reward: UMaze, SimpleCrossing, LavaCrossing, DoorKey, and ObjectDelivery. In U-Maze, the agent must traverse an indirect path around a wall to the goal, in a 
``U'' shaped maze on a 5x5 grid. SimpleCrossing and LavaCrossing are goal-oriented navigation tasks in a 9x9 grid. The agent begins in the top left and must reach the goal in the bottom right, navigating around procedurally generated walls (or lava). In the DoorKey environment, a locked door divides the arena into two rooms. The agent must reach the goal in the other room by fetching a key. Lastly, in the ObjectDelivery task family, three objects of type \emph{ball} or \emph{box} and color \emph{blue} or \emph{red} are randomly placed. One object is randomly selected to be transported to rest at a goal location. There are 107,520 possible configurations of objects.

\paragraph{Abstractions}

\begin{wrapfigure}{r}{0.5\linewidth}
  \begin{center}
  \vspace{-5pt}
    \includegraphics[width=0.9\linewidth]{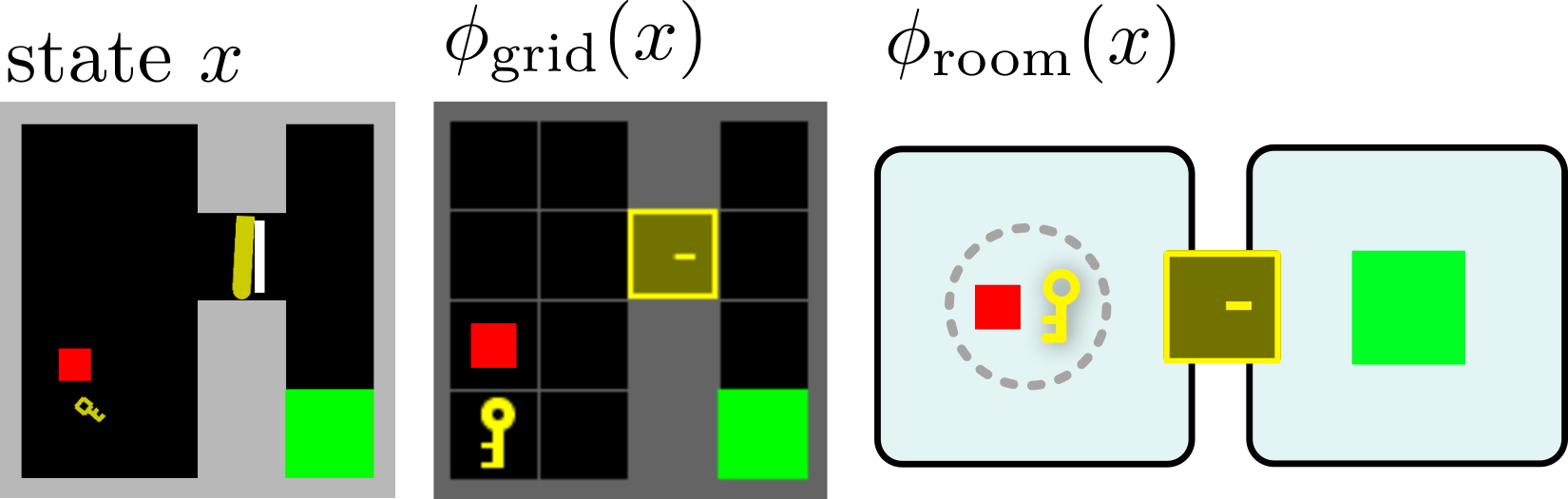}
  \end{center}
  \caption{Visualization of the ``grid'' and ``room'' abstractions. In the continuous state $x$, the red agent has grabbed a yellow key. In the grid abstraction, the agent is on a grid cell with the key. In the room abstraction, two rooms are separated by a yellow door. The agent and key are near each other (dotted circle) but not near the door.}
  \label{fig:abstractions}
  \vspace{-3pt}
\end{wrapfigure}
To test how expert domain knowledge helps agents adapt in new environments, we construct two discrete abstractions for the CocoGrid environments: a grid abstraction $\langle \phigrid, \phi_\mathrm{grid} \rangle$ and a room abstraction $\langle \phiroom, \phi_\mathrm{room} \rangle$, each providing different levels of detail to the agent (Figure \ref{fig:abstractions}). The abstractions are reasonable descriptions of the environment a human might provide about relevant state features and transitions.
$\phigrid$ captures the full structure of the original Minigrid environment. 
Positions are discretized into grid cells with objects characterized by type and color. The agent can move to adjacent cells. $\phiroom$ instead aggregates contiguous empty grid cells into ``rooms" separated by doors. Precise locations within rooms are replaced with a set of ``near" relations. The agent can move near an object in the same room or travel through a door to another room.

Some caution must be employed when designing abstractions. Since doors do not physically occupy an entire grid cell, a naive abstraction would hide which side of a locked door the agent is on, causing the planner to give a subgoal that is impossible to achieve. This is an example of \emph{aliasing} \citep{zhang_composable_2018}. We correct this by assigning the region on either side of a door to its neighbor cell. 

Still, the abstractions do not faithfully represent the continuous dynamics. Agents can slip through a partially-opened door before the abstraction considers it open. A held object can cross into another grid cell before the agent does. An effective controller should tolerate some degree of inaccuracies in the abstractions and even improve upon the high level plan.

\section{Results and Discussion}

\begin{wrapfigure}{r}{0.35\linewidth}
  \begin{center}
  \vspace{-5pt}
\includegraphics[scale=0.35]{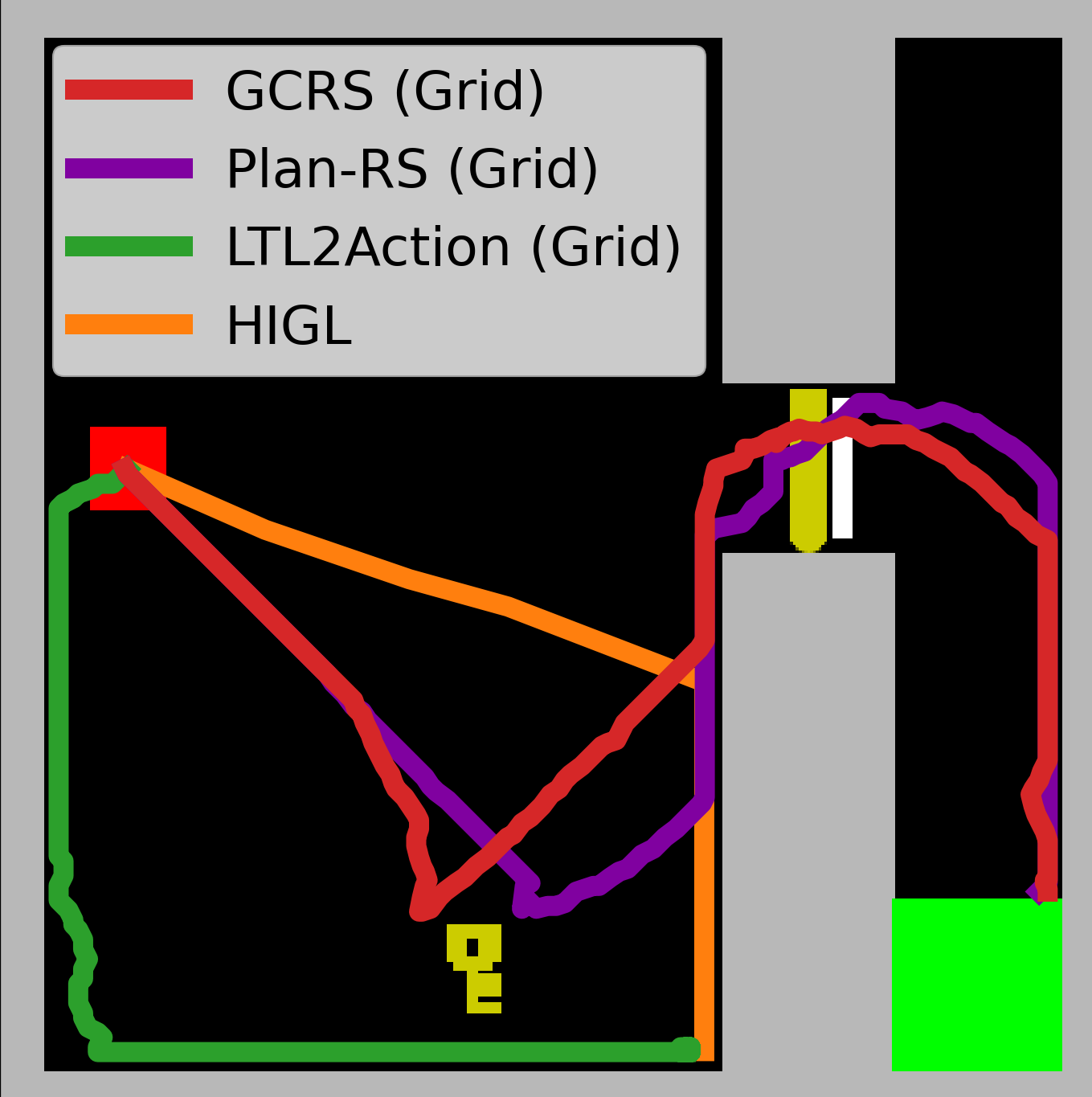}
  \end{center}
  \caption{Visualizing trajectories on DoorKey-8x8. GCRS and Plan-RS with $\phigrid$ were successful.}
  \label{fig:trajectory}
  \vspace{-3pt}
\end{wrapfigure}

\begin{figure*}[t]
\centering

\includegraphics[width=0.6\linewidth]{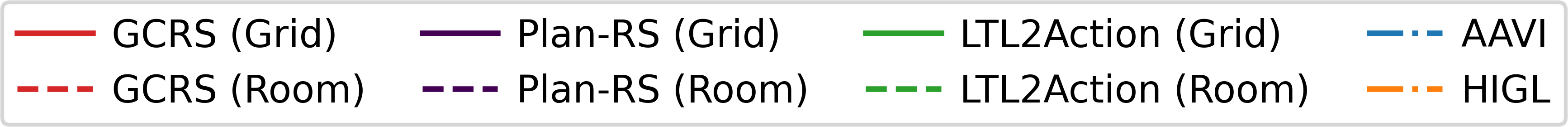}
  \includegraphics[width=0.8\linewidth]{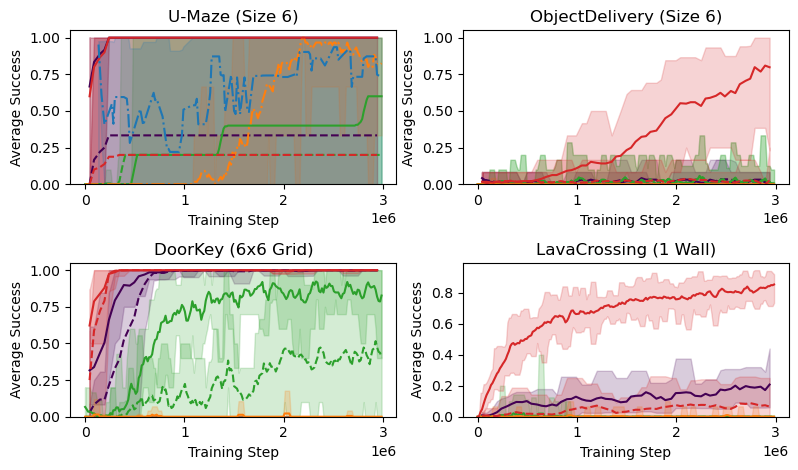}
  \captionof{figure}{Average success rates and standard deviation of different techniques, over 5 training runs.}
  \label{fig:train_success}
\end{figure*}

\paragraph{Sample Efficiency} Figure \ref{fig:train_success} shows the task success rate averaged over five training runs during training on U-Maze, ObjectDelivery, DoorKey, and LavaCrossing. GCRS with $\phigrid$ consistently has the highest success rate in fewer samples. GCRS solved DoorKey efficiently with $\phiroom$. However, for U-Maze and LavaCrossing, $\phiroom$ has only one room, so $\mH^\tau$ gives extremely sparse feedback, resulting in the poor performance. Plan-RS matched the performance of GCRS for U-Maze and close behind for DoorKey, yet struggled with the more randomized tasks. LTL2Action was eventually successful at DoorKey with both the abstractions, and modestly successful with the U-Maze. HIGL was unsuccessful on these tasks, as it seemed unable to handle changing environments. AAVI had very high variance on the U-Maze, owing to it getting stuck on corners at the seam between options. Furthermore, AAVI could not even train on the procedural environments because they had many abstract states that required too many options to fit into memory.

\begin{figure*}[t]
\centering

\includegraphics[width=0.6\linewidth]{legend_4x2_crop.png}
\includegraphics[width=0.77\linewidth]{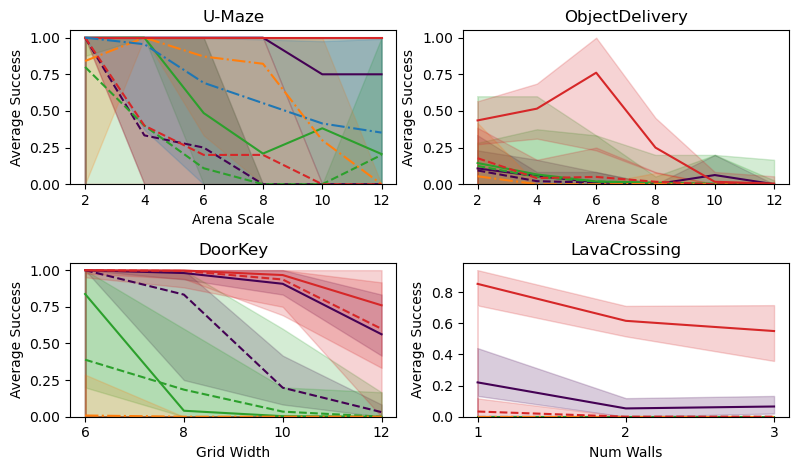}
  \captionof{figure}{Effect of scaling environment difficulty level on task completion rate. For U-Maze and ObjectDelivery, scale multiplies the physical size of the arena. 
 DoorKey and LavaCrossing are scaled against the width of the grid and the number of lava walls respectively. }
\label{fig:scaling_success}
\end{figure*}

\paragraph{Scaling Environment Difficulty} We evaluated the success rate on U-Maze and ObjectDelivery as the size of the arena was scaled between $2$-$12$ (Figure~\ref{fig:scaling_success}). The size scales \emph{linearly} with the number of steps required to reach the goal, affecting the reward sparsity. GCRS and Plan-RS with the grid abstraction solved U-Maze tasks on nearly all scales.
In ObjectDelivery, GCRS outperformed other methods until scale 10. For the DoorKey domain, we scaled the number of arena grid cells from $6$x$6$ to $12$x$12$. This increases the number of possible locations for the key and door, and requires longer plans when using the grid abstraction. GCRS maintained near-perfect success on the grid and room abstractions until $10$x$10$ before dropping off slightly. Plan-RS and LTL2Action were successful in smaller grids, but their performance degraded quickly as difficulty increased.
In LavaCrossing, GCRS with $\phigrid$ was significantly better than other methods, even as the number of walls increased. $\phigrid$ also helped Plan-RS, but despite the same reward structure, without subgoals it could not adapt to procedurally generated hazards. LTL and HIGL failed with even one wall in this domain. AAVI still did not run on the procedural environments.

\begin{figure}[t!]
    \begin{subfigure}{0.3\linewidth}\centering
    \includegraphics[scale=0.4]{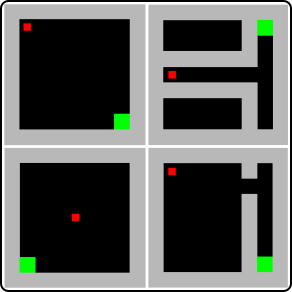}
\subcaption{}\end{subfigure}
\begin{subfigure}{0.2\linewidth}\centering\includegraphics[scale=0.4]{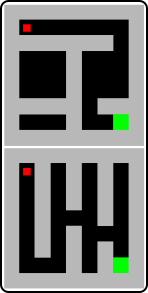}
\subcaption{}\end{subfigure}\hfill
    \begin{subfigure}{0.5\linewidth}\centering \includegraphics[scale=0.06]{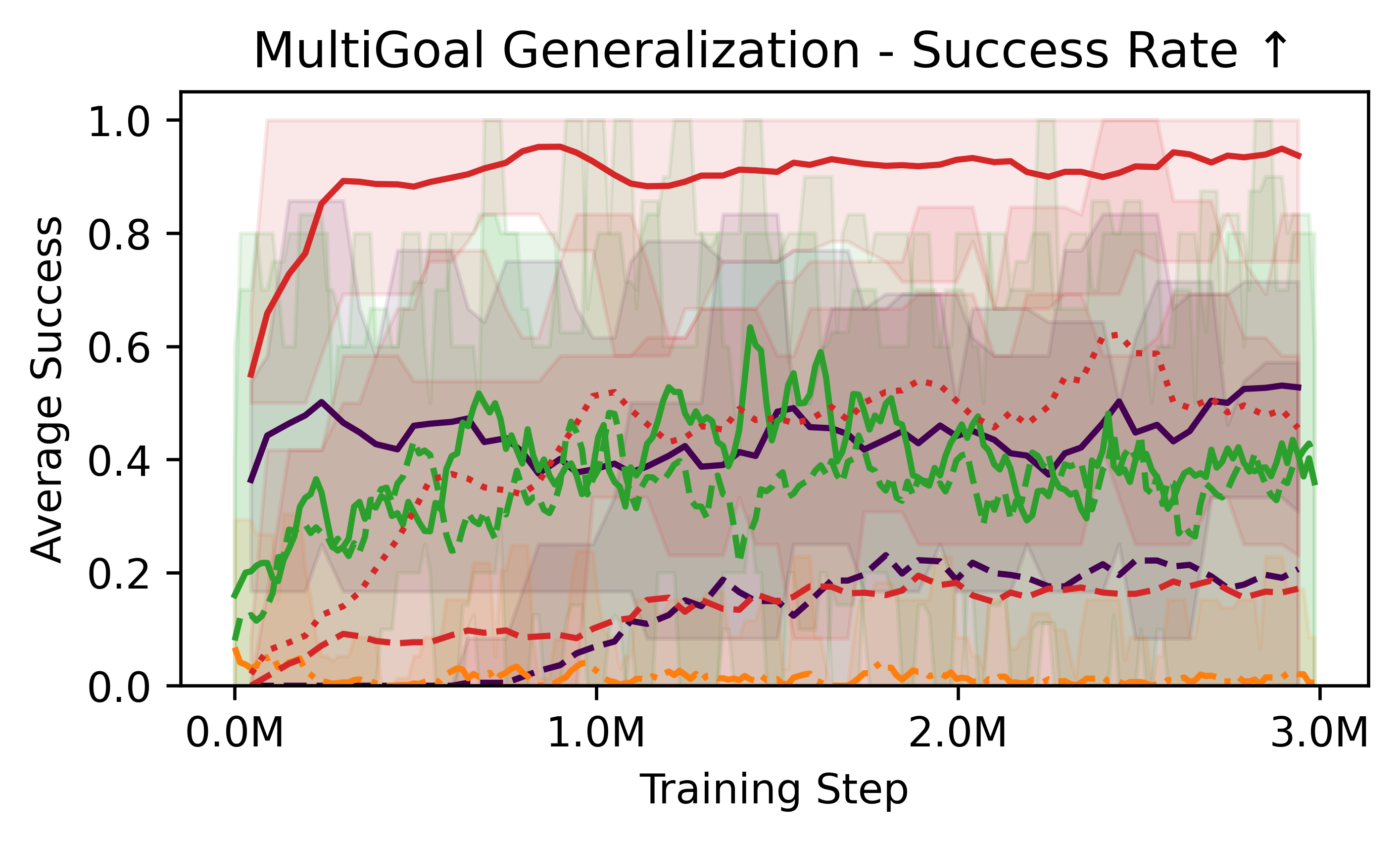}
\subcaption{}\end{subfigure}

    \begin{subfigure}{0.15\linewidth}\centering\includegraphics[scale=0.4]{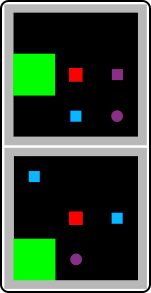}
        \subcaption{}\end{subfigure}
        \begin{subfigure}{0.15\linewidth}\centering\includegraphics[scale=0.4]{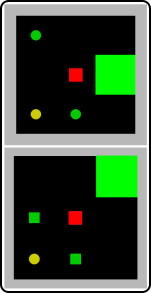}
        \subcaption{}\end{subfigure}\hfill
    \begin{subfigure}{0.18\linewidth}
    \centering
    \includegraphics[width=\linewidth]{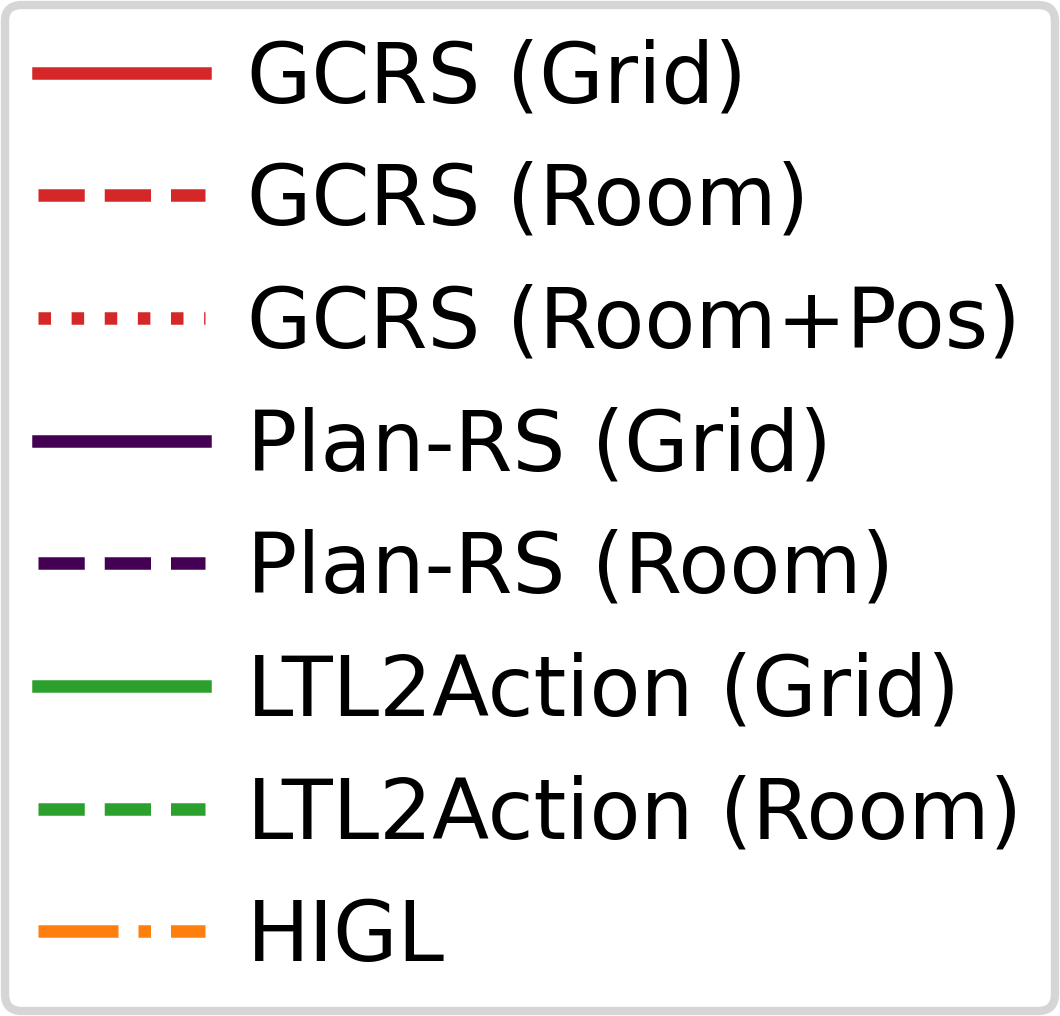}
    \subcaption*{}
    \end{subfigure}
    \begin{subfigure}{0.48\linewidth}\centering \includegraphics[scale=0.06]{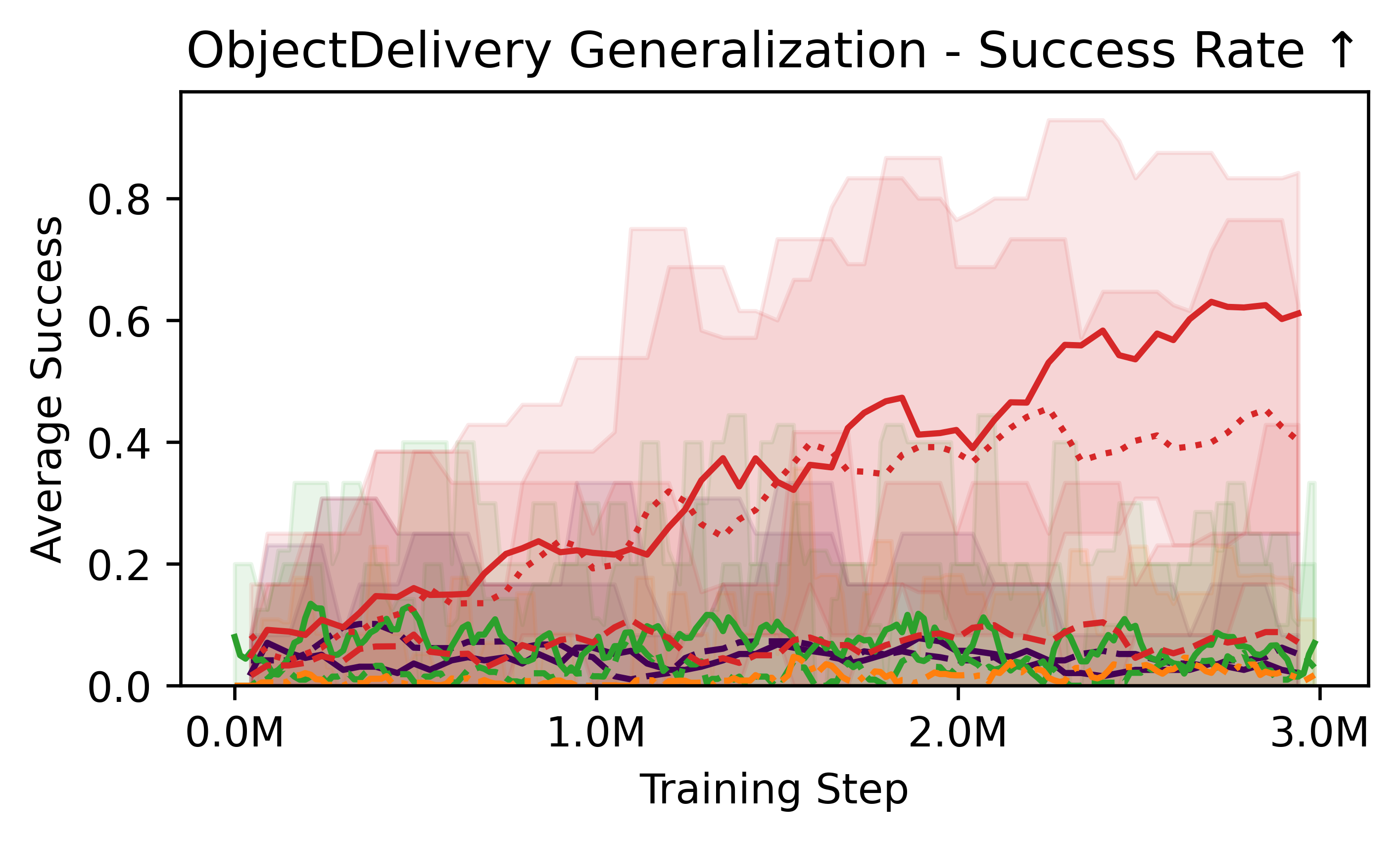}
        \subcaption{}\end{subfigure}
    \caption{Zero-shot generalization. (a) MultiGoal environment training distribution illustration, with instances from Empty, HallwayChoice, RandomCorner, and SimpleCrossing-1-Wall. (b) Illustration of evaluation instances from SimpleCrossing-2-Wall and SimpleCrossing-3-Wall. SimpleCrossing is procedurally generated each episode. (c) Average task completion rate on evaluation instances with no additional training. (d) Illustration of ObjectDelivery training with red and blue objects. (e) Illustration of evaluation tasks where the agent must transport green or yellow objects to the goal. (f) Average task completion rate in ObjectDelivery evaluation instances with no additional training. }
        \label{fig:multigoal}
\end{figure}

\paragraph{Zero-Shot Generalization}
We test the generalization capabilities of our learned controller on a suite of procedurally generated mazes, referred to as MultiGoal, and ObjectDelivery with different object colors from training (Figure~\ref{fig:multigoal}). In MultiGoal, we train the agent using several maze configurations that were uniformly sampled, including SimpleCrossing with one wall (Figure \ref{fig:multigoal}(a)). The evaluation tasks are sampled from SimpleCrossing with two and three walls (Figure \ref{fig:multigoal}(b)), requiring the agent to change directions multiple times.
Figure \ref{fig:multigoal}(c) shows average success rate of different techniques on the evaluation distribution over the course of training. GCRS with $\phigrid$ consistently achieves over $90\%$ success on the harder environments. In the ObjectDelivery domain, agents are trained on red and blue objects (Figure \ref{fig:multigoal}(d)), but evaluated on green and yellow objects (Figure \ref{fig:multigoal}(e)). 
Note that the agents are trained with vector observations, with colors given as ordinal identifiers. To succeed, the color in the task must be cross-referenced with the corresponding object position. Similar to MultiGoal, we observe that GCRS with $\phigrid$ generalizes well, achieving $60\%$ success on green and yellow objects (Figure~ \ref{fig:multigoal}(f)). This generalization is explained by the native systematicity~\citep{kirk_survey_2023} of the grid abstraction, applying rules to plan over new states. While the network policy has no experience with green or yellow, $\phigrid$ provides the policy a subgoal position similar to training. GCRS with $\phiroom$ fails on the new colors, but we ran it again with the position of the target object in the subgoal encoding, $\mH_\mathrm{room+pos}$. This significantly improved over $\phiroom$, demonstrating that even a sparser abstraction can be beneficial. Overall, our results show that dynamic subgoals provide a strong guidance for zero-shot generalization and the performance is even better with more informative abstractions such as $\phigrid$. 

\section{Summary}
We present a hierarchical RL approach that uses an expert-defined abstraction for efficient learning in continuous environments with sparse rewards. Our approach dynamically plans over expert-defined abstractions to generate subgoals that help the agent learn a goal-conditioned policy, using plan-based reward shaping. Our empirical evaluation demonstrates the sample efficiency, scalability, and zero-shot generalization capabilities of our approach. Our results show that reward shaping boosts training speed and success, and the high-level planner's guidance gives a significant advantage in completing tasks in procedural environments.
In the future, we aim to investigate techniques for adaptive abstraction refinement that augments information to the expert-provided abstractions based on agent experience. We also aim to extend this work to multi-agent settings where the agents will benefit from a global abstraction that simultaneously guides the learning of multiple agents. 

\subsubsection*{Acknowledgments}
\label{sec:ack}

This work was supported in part by DARPA TIAMAT HR0011-24-9-0423.

\bibliography{main}
\bibliographystyle{rlj}

\appendix

\section{Proof of Proposition 1}
\label{sec:appendix1}

\begin{proof}
Consider an MDP $\mM' \doteq \langle \mX', \mU, p', R', \mu_0', \gamma \rangle$, with state space $\mX' \doteq \mX \times S \times\mT$, reward $R[(x, s, \tau), u, (x', s', \tau')] \doteq \mathds{1}[\tau = \tau']R^\tau(x, u, x')$, initial state distribution $\mu_0(x, s, \tau) \doteq \mu_\tau(\tau)\mu_0^\tau(x)$, and dynamics $ p'[(x, s, \tau), u, (x', s', \tau')] \doteq p(x, u, x') $ if $\tau = \tau', s = \mathop{subgoal}(x, \tau)$ and $s' = \mathop{subgoal}(x', \tau)$, and $p'[(x, s, \tau), u, (x', s', \tau')] \doteq 0$ otherwise. 
The objective for $\mM'$ is to find a $\pi' : \mX' \to \Delta(U)$ that maximizes 
\begin{align}
    J'(\pi') =&\, 
    \mathbb{E}_{(x_0, s_0, \tau) \sim \mu_0} \left[ \mathbb{E}_{\pi'} \left[\sum_t \gamma^t R'[(x,s,\tau), u, (x', s', \tau')] \middle| x_0 \right]\right].
\end{align}

Since for every transition, the task remains fixed, the reward simplifies to $R'((x,s,\tau), u, (x', s', \tau')) = R^\tau(x,u,x')$. Because the initial state $(x_0, s_0, \tau)$ is drawn from $\mu_0(x, \tau)=\mu_\tau(\tau)\mu_0^\tau(x)$, we can write
\begin{align}
    J'(\pi') =&\,
    \mathbb{E}_{\tau \sim \mu_\mT} \left[ \mathbb{E}_{x_0 \sim \mu_0^\tau} \left[ \mathbb{E}_{\pi'} \left[\sum_t \gamma^t R^\tau(x, u, x') \middle| x_0 \right]\right]\right].
\end{align}

This is identical to $J(\pi)$, except $\pi$ takes an input $\mX \times \mT$, not $\mX \times S \times \mT$. Let $\pi(x, \tau) \doteq \pi'(x, \mathop{subgoal}(x, \tau), \tau)$ be a policy in $\mM$. Then $J(\pi)=J'(\pi')$. Since $\mathop{subgoal}(x,\tau)$ is a deterministic mapping, there is a one-to-one correspondence between policies in $\mM$ and $\mM'$. Therefore, if $\pi'$ is optimal in $\mM'$, then $\pi$ is optimal in $\mM$.

Now consider another MDP $\mM_\mathrm{shaped}'$, identical to $\mM'$, but with potential-based reward shaping (PBRS), $R[(x, s, \tau), u, (x', s', \tau')] = R^\tau(x, u, x') + \gamma \Phi^\tau(x') - \Phi^\tau(x)$, where $\Phi^\tau(x) = V_{H^\tau}^*(\phi(x))$.
Suppose the policy update on Line~\ref{line:update_policy} is an RL procedure that converges to an optimal policy under its usual assumptions.
Then Algorithm \ref{alg:rl_amdp} gives the optimal policy for $\mM_\mathrm{shaped}'$, denoted by $\pi'^{*}$. According to \citet{canonaco_sample_2024}, in goal-oriented episodic MDPs, the optimal policy from the unshaped MDP is preserved by PBRS, so $\pi'^{*}$ is optimal in $\mM'$.
Consequently, $\pi^*(x, \tau) \doteq \pi'^*(x, \mathop{subgoal}(x, \tau), \tau)$ is optimal for $\mM$.
\end{proof}


\end{document}